\documentclass{article}
\usepackage{ijcai16}

\usepackage{times}
\usepackage{color}
\usepackage{url}
\usepackage{multirow}
\usepackage{rotating}
\usepackage{bbm}
\usepackage{latexsym}
\usepackage{amsthm}
\usepackage{amsmath}
\usepackage{float}
\usepackage{mathtools}
\usepackage{accents}
\usepackage{bbm}
\usepackage{enumerate}
\usepackage{afterpage}
\usepackage{graphicx}
\usepackage{amssymb}
\usepackage{epsfig}
\usepackage[ruled,lined,boxed]{algorithm2e}

\newcommand{\cS}{{\mathcal S}}
\newcommand{\cA}{{\mathcal A}}

\newcommand{\cU}{{\mathcal U}}

\def\app#1#2{%
  \mathrel{%
    \setbox0=\hbox{$#1\sim$}%
    \setbox2=\hbox{%
      \rlap{\hbox{$#1\propto$}}%
      \lower1.1\ht0\box0%
    }%
    \raise0.25\ht2\box2%
  }%
}

\newcommand{\suchthat}{\;\ifnum\currentgrouptype=16 \middle\fi|\;}
\newcommand{\dotprod}[1]{\langle #1 \rangle}

\newcommand{\eg}{\textit{e.\@g.\@}}
\newcommand{\ie}{\textit{i.\@e.\@}}

\DeclareMathOperator{\E}{\mathbb{E}}
\DeclareMathOperator{\N}{\mathbb{N}}
\DeclareMathOperator{\R}{\mathbb{R}}

\SetKwComment{dbc}{//}{}
\SetFuncSty{sc}
\LinesNumbered
\DontPrintSemicolon
\SetAlCapNameSty{sc}
\SetProcNameSty{sc}

\SetKwFunction{init}{Initialize}
\SetKwFunction{rand}{Random}
\SetKwFunction{cdmisfa}{Curious Dr. MISFA}
\SetKwFunction{oselect}{Option Selection}
\SetKwFunction{clip}{Clip}
\SetKwFunction{mlspi}{Model-LSPI}

\title{Intrinsically Motivated Acquisition of Modular Slow Features \\ for Humanoids in
Continuous and Non-Stationary Environments}
\author{Varun Raj Kompella and Laurenz Wiskott \\
Institute for Neural Computation, Ruhr-Universitat Bochum\\
\{varun.kompella, laurenz.wiskott\}@ini.rub.de}

\begin{document}

\maketitle

\begin{abstract}

A compact information-rich representation of the environment, also called a feature abstraction, can simplify a robot's task of mapping its raw sensory inputs to useful action sequences. However, in environments that are non-stationary and only partially observable, a single abstraction is probably not sufficient to encode most variations. Therefore, learning multiple sets of spatially or temporally local, modular abstractions of the inputs would be beneficial. How can a robot learn these local abstractions without a teacher? More specifically, how can it decide from where and when to start learning a new abstraction? A recently proposed algorithm called Curious Dr.\ MISFA addresses this problem. The algorithm is based on two underlying learning principles called \emph{artificial curiosity} and \emph{slowness}. The former is used to make the robot self-motivated to explore by rewarding itself whenever it makes progress learning an abstraction;   the later is used to update the abstraction by extracting slowly varying components from raw sensory inputs. Curious Dr.\ MISFA's application is, however, limited to discrete domains constrained by a pre-defined state space and has design limitations that make it unstable in certain situations. This paper presents a significant improvement that is applicable to continuous environments, is computationally less expensive, simpler to use with fewer hyper parameters, and stable in certain non-stationary environments. We demonstrate the efficacy and stability of our method in a vision-based robot simulator.

%

\end{abstract}

\section{Introduction}

Reinforcement learning~(RL)~\cite{Kaelbling:96,Sutton:98} provides a basic framework for an actively exploring agent to acquire desired task-specific behaviors by maximizing the accumulation of task-dependent external rewards through simple trial-and-error interactions with the environment. In high-dimensional real world environments, however, RL can be slow since external rewards are usually sparsely available and can sometimes be extremely difficult to obtain by pure random exploration. Fortunately, most real world transitions lie on a low-dimensional manifold. Learning a compact representation (feature abstraction) of the environment sensed through high-dimensional sensory inputs, can therefore speed up exploration and the subsequent task learning~\cite{lange2010deep,legenstein2010reinforcement,AutoIncSFA2011,koutnik2014evolving,mnih2015human}. 

In environments that are non-stationary and partially observable, a single abstraction is probably not sufficient to encode most variations, in which case it would be beneficial to learn a repertoire of spatially or temporally local abstractions that can potentially be translated to multiple skills. In the absence of external supervision, how can the agent be motivated to learn these abstractions? The agent would need to be intrinsically motivated. Over the recent years, intrinsic motivation (IM) has been considered a useful tool for adaptive autonomous agents or robots~\cite{schmidhuber2013maximizing,baldassarre2013intrinsically}. There exists several computational approaches that model different IM signals for RL agents, for example, IM signals that are based on novelty~\cite{itti:05}, prediction error~\cite{Schmidhuber:90sab,Barto:04}, knowledge/prediction improvements~\cite{Schmidhuber:91singaporecur} and those that are based on the competence to reach a certain goal~\cite{SchembriMirolliBaldassarre2007EvolutionLearningIntrinsicallyMotivated1111459278}. Refer to~\cite{schmidhuber2013maximizing,baldassarre2013intrinsically} for a survey on the pros and cons of these approaches. 

Most of the intrinsically motivated RL techniques have been applied to exploring agents in simple domains~\cite{Bakker:04ias,stout2010competence,pape2012learning,santucci2015best}, agents that use hand-designed or pre-trained state abstractions of high-dimensional environments~\cite{konidaris2011autonomous,ngo2013Frontiers}, or agents that are provided with a low-dimensional task-space~\cite{baranes2013active}. Very few have addressed the issue of learning task-independent low-dimensional abstractions from high-dimensional inputs while simultaneously exploring the environment. The main problem in such scenarios is to learn  abstractions from non-\emph{i.i.d} and potentially non-stationary sensory inputs that are a function of the agent's actions and other unknown time-varying factors in the environment.  Mugan and Kuipers QLAP~\cite{mugan2012autonomous}, Xu and Kuipers OSH~\cite{Xu-PhD-11} and Kompella et al.'s Curious Dr.\ MISFA~\cite{luciw2013intrinsic,kompellaThesis} are a few closely related examples in the direction of learning feature abstractions from action sequences that are specific to localized regions in the environment. QLAP learns simplified predictable knowledge by discretizing low-level sensorimotor experience through defining landmarks and observing contingencies between the landmarks. It assumes that there exists a low-level sensory model that can, e.g., track the positions of the objects in the scene. OSH builds a collection of multi-level object representations from camera images. It uses a ``model-learning through tracking''~\cite{modayil2008initial} strategy to model the static background and the individual foreground objects assuming that the image background is static. 

Curious Dr.\ MISFA is by far the closest that comes to addressing the problem of learning task-independent multiple abstractions from raw images online in the absence of any external guidance. The agent actively explores within a set of high-dimensional video streams\footnote{A video stream could be generated as a consequence of executing a particular agent's behavior.} and learns to select the stream where it can find the next easiest (quickest) to learn a \emph{slow feature} (SF;~\cite{WisSej2002}) abstraction. It does this \emph{optimally} while simultaneously updating the SF abstractions using Incremental Slow Feature Analysis (IncSFA;~\cite{kompella:nc2012}). IncSFA is based on the \emph{slowness principle}~\cite{mitchison1991removing,Foldiak:95}, which
states that the underlying causes of fast changing inputs vary at a much 
slower timescale. IncSFA uses the temporal correlations within the inputs to extract SFs online. SFs have been shown to be useful for RL as they capture the transition process generating the raw sensory inputs~\cite{wiskott2003estimating,sprekeler2011relation,AutoIncSFA2011,luciw:icann2012,bohmer2013construction}. The result of the learning process of Curious Dr.\ MISFA is an optimal sequence of SF abstractions acquired in the order from easy to difficult-to-learn ones, principally similar to the learning process of Utgoff and Stracuzzi's many-layered learning~\cite{utgoff2002many}. Curious Dr.\ MISFA has also been used to show a continual emergence of reusable unsupervised skills on a humanoid robot (topple, grasp, pick-place a cup) while acquiring SF abstractions from raw-pixel vision~\cite{kompella14ijcnn,kompella2015continual}, the first of its kind. 

Curious Dr.\ MISFA's application is, however, limited to discrete domains constrained by a pre-defined discrete state space and has design limitations that make it unstable in certain situations. This paper presents a significant improvement that is applicable to continuous environments, is computationally less expensive, simpler to use with fewer hyper parameters, and stable in non-stationary environments where the statistics change abruptly over time. We demonstrate these improvements empirically and make our Python code of the algorithm available online as open source. Next, we discuss details of our proposed algorithm.

\section{CD-MISFA 2.0}
\label{SE:OVRW}

We discuss here the details of our new method. To keep it short, we refer to the original Curious Dr.\ MISFA as CD-MISFA 1.0 and our new method as CD-MISFA 2.0 (refer Section \ref{SE:RESULTS_COMP} for a detailed comparison between the two methods). Next, we provide an intuitive analogical example to explain the underlying problem that is being solved.  

\textbf{Intuition.} Consider a camera equipped agent viewing different channels on a television. Each channel generates a continuous stream of images (that may or may not be predictable). The agent at any instant can access information only from a single channel. It can explore the channels by selecting a particular channel for a period of time and then switch. The distribution of images received by the agent as a consequence of its exploration, in most cases, is non-stationary. This makes it infeasible to learn a single abstraction encoding all the channel streams. The problem can be simplified by learning abstractions of individual channels that generate inputs from a stationary distribution. But how can the agent find out (a) the channel and (b) for how long to observe the channel, to know that there exists a stationary distribution? We discuss next the details of the CD-MISFA 2.0 algorithm that addresses a general version of this problem.

\textbf{Environment.} The environment considered is similar to the one of CD-MISFA 1.0. It consists of \emph{n} sources of observation streams $X = \{{\bf x}_1, ..., {\bf x}_n: {\bf x}_i(t) = (x_i^1(t),..., $ $x_i^I(t))\in \R^{I\in \N}\}$. These streams could be image sequences observed over different head rotation angles of a robot or while executing different time-varying behaviors. The agent explores the streams with two actions: \{stay, switch\}. When the agent takes the \emph{stay} action, the current stream ${\bf x}_i$ remains the same and it receives a hand-set number of $\tau$ observations from that stream. When it takes the action \emph{switch}, the agent selects a stream ${\bf x}_{j \neq i}$ uniformly randomly from one of the other $n-1$ streams and it receives $\tau$ observations from the new stream. 

\textbf{Goal.} The goal of the agent is to learn a sequence of slow feature abstractions $\Phi = \{\phi_1,...,\phi_m;$ $m \le n\}$ that each encode one or more of the observation streams in $X$. $\phi_i$ is generally a matrix of parameters. The order of the sequence is such that $\phi_1$ encodes the easiest and $\phi_m$ the most difficult learnable stream in $X$. CD-MISFA 2.0 achieves this goal by iterating over the following steps: (1) Find the easiest novel observation stream while simultaneously learning an abstraction encoding it. (2) Store the abstraction and use it to filter known or similar observation streams. (3) Continue with step (1) on the remaining streams.

\textbf{Architecture.} The architecture includes: 

\textit{(a) Adaptive abstraction.} A single adaptive abstraction $\widehat{\phi}$ is updated online using IncSFA for each observation ${\bf x}(t)$. Details on the learning rules of IncSFA can be found in Kompella's previous work~\cite{kompella:nc2012}. The instantaneous output of the adaptive abstraction for the observation ${\bf x}(t)$ is given by:
\begin{align}
\label{EQ:INSTFEATOUT}
 \quad {\bf y}(t) = \widehat{\phi}{\bf x}(t).
\end{align}


\textit{(b) Gating system.} A gating system is used to accomplish two tasks: (1) Decide when to stop updating $\widehat{\phi}$ and store it $\phi_i \leftarrow \widehat{\phi}$. Once stored, $\phi_i$ is frozen and a new $\widehat{\phi}$ is created. (2) Use the stored frozen abstractions to filter observations from known or similar input streams while updating the new $\widehat{\phi}$. 

For the first task, we estimate and use the time derivative of the \emph{slowness measure}~\cite{WisSej2002}. Slowness measure of a time-varying signal $y$ is defined as:
\begin{align}
\eta(y) = \frac{1}{2~\pi}  \sqrt{\frac{\E(\dot{y}^2)}{\text{Var}(y)}},
\end{align}
where $\dot{y}$ represents the temporal derivative of $y$, $\E$ and $\text{Var}$ represent the expectation and variance. This measure quantifies how fast or slow a signal changes in time. We compute $\eta$ values of all the output components of the adaptive abstraction online. When the abstraction has converged, the $\eta$s will converge as well and their derivative will tend towards zero. The gating system uses the following condition to check when to stop updating the adaptive abstraction:
\begin{align}
\label{EQ:ETADERIVDELTA}
 |\dot{\eta}(y^i(t)) | < \delta, \forall y^i(t) \in {\bf y}(t).
\end{align}

For the second task, we compute an instantaneous $\eta$
\begin{align}
\label{EQ:INSTETA}
\eta^{\text{inst}}(y) = \frac{1}{2~\pi}  \sqrt{\frac{\E_{\tau}(\dot y^2)}{\text{Var}_{\tau}( y)}},
\end{align}
for each output component $y^i \in {\bf y}$, where $\E_{\tau}$ and $\text{Var}_{\tau}$ are the mean and variance of only the $\tau$ samples. When $\tau$ is large $\eta^{\text{inst}}(y) = \eta(y)$. We also track a moving standard deviation (SD) for each $\eta^{\text{inst}}(y^i(t))$. When $\widehat{\phi}$ is saved, the estimated SDs are also saved. To find out if a new set of $\tau$ samples is novel, $\eta^{\text{inst}}$ of all the frozen abstractions are computed for the new samples according to Eq.~\eqref{EQ:INSTETA} and then checked if they lie outside two times their corresponding SDs. 

\textit{(c) Curiosity-Driven Reinforcement Learner (CDRL).} A CDRL is used to find (a) the unknown order of the observation streams in terms of the difficulty of learning them with IncSFA, and (b) the optimal sequence of actions (\emph{stay} or \emph{switch}) required to learn $\Phi$. Let $s \in \cS = \{s_1, ..., s_n\}$ denote the indices of the observation streams and $u \in \cU = \{u_1, ..., u_m\}$ denote the indices of the abstractions to be learned. Let $\cA = \{0=\text{stay},1= \text{switch}\}$. The goal of the CDRL reduces to learning an observation stream selection policy $\pi^* : \cS \times \cU \rightarrow \cA$ that maps an optimal action for each stream ${\bf x}_i$ to learn the abstraction $\phi_i$. For example, consider an environment with $5$ streams with ${\bf x}_3$ being the easiest to learn and ${\bf x}_1$ the next. To learn the first abstraction, $\pi^*(.,u_1)$ is a vector $[1,1,0,1,1]$, and the second abstraction $\pi^*(., u_2)$ is $[0,1,1,1,1]$. How can the CDRL learn such a $\pi^*$? Since the desired $\Phi$ is an ordered finite set of unique abstractions, it follows that the corresponding sub-policy $\pi^*(.,u_i)$ (denoted in short as $\pi^*_{u_i}$) required to learn the abstraction $\phi_i$ is unique. Therefore, $\pi^*$ is learned sequentially by learning unique sub-policies in the order $\{\pi^*_{u_1}, ..., \pi^*_{u_m}\}$. 

The convergence of the agent's sub-policies $\pi_{u_i} : \cS \rightarrow \cA$ to their optimal ($\pi_{u_i}^*$) is guided through internal rewards for each tuple ($\text{current state } s, \text{ current action } a, \text{ future state }s'$): 
\begin{align} \label{EQ:REWPROG}
r^{ss'}_a = \left(-\dotprod{\dot{\xi}}_t^{\tau} + \beta Z^{\sigma}(\dotprod{\xi}_t^{\tau})\right), 
\end{align}
where $Z^{\sigma}(x) = \mathrm{e}^{-x^2 / 2\sigma^2}$ is a Gaussian function, $\sigma$ and $\beta$ are scalar constants, $\xi(t)$ denotes the Frobenius norm $\|\widehat{\phi}(t+1) - \widehat{\phi}(t)\|$ and $\dotprod{\xi}_t^{\tau} = \frac{1}{\tau} \sum^{t+\tau-1}_t \xi(t), ~\dotprod{\dot{\xi}}_t^{\tau} = \dotprod{\xi}_t^{\tau} - \dotprod{\xi}_{t-\tau}^{\tau}$.
The RL objective learns a policy that maximizes the accumulation of these rewards over time. There are two terms in the reward equation, maximizing the first term would result in a policy that shifts the agent to states where the weight-change decreases sharply ($\dotprod{\dot{\xi}}_t^{\tau} < 0 $). This term is often referred in the literature as the \emph{curiosity} reward~\cite{schmidhuber2006developmental,schmidhuber2010formal}. Intuitively, the curiosity reward term is responsible for finding the easiest observation stream. Maximizing the second objective results in a policy that improves the developing $\widehat{\phi}$ to better encode the observations, making it an expert. We refer to the second term as the \emph{expert} reward. 

A reward function $R:\cS \times \cA \times \cS \rightarrow \R$ (tensor of size $|\cS|\times|\cA|\times|\cS|$) is estimated online using the instantaneous rewards as:
\begin{align}
\label{EQ:REWARDUPDATE}
 R \leftarrow \frac{1}{t} \widetilde{R} + \left(1 - \frac{1}{t}\right)~R,
\end{align}
where $\widetilde{R}$ is an instantaneous tensor (same shape as $R$) with its $(s,a,s')$ component equal to the instantaneous reward $r^{ss'}_a$ and all other components equal to zero. After every $\tau$ observations, a value function $Q$ and the sub-policy $\pi_{u_i}$ are updated using the estimated $R$ via Least Squares Policy Iteration (LSPI;~\cite{lagoudakis2003least}).

\begin{figure}[!t]
\begin{center}
\includegraphics[width=\linewidth]{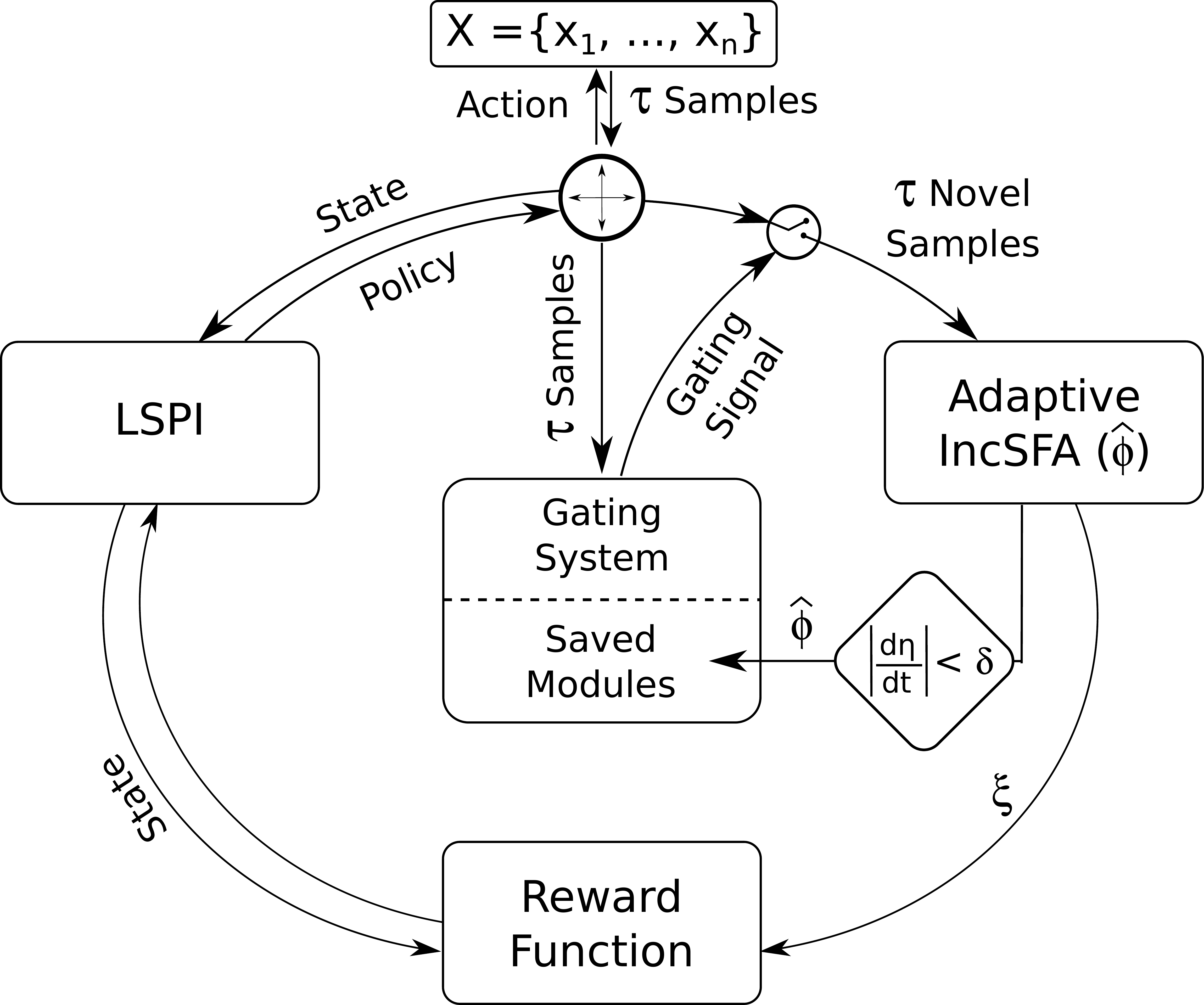}
\caption{Control flow diagram of CD-MISFA 2.0 algorithm. }
\label{FIG:Flow_Comp}
\end{center}
\vspace{-0.3cm}
\end{figure}

\textbf{Learning Process.} Figure~\ref{FIG:Flow_Comp}(a) shows the control flow diagram of the algorithm. At $t=0$, the agent begins by observing $\tau$ samples from the current stream. Since there are no previously learned abstractions, the set of $\tau$ samples is novel and is used to update $\widehat{\phi}$. Condition~\eqref{EQ:ETADERIVDELTA} is checked and if not met, $R$ and $\pi_{u_1}$ are updated according to Eq.~\eqref{EQ:REWARDUPDATE}  and LSPI algorithm respectively. The agent uses a decaying $\epsilon$-greedy strategy~\cite{Sutton:98} on $\pi_{u_1}$ to take a new action and the process repeats. After a few iterations, the sub-policy $\pi_{u_1}$ converges to $\pi_{u_1}^*$. The converging $\pi_{u_1}$ also enables $\widehat{\phi}$ to converge. When $|\dot{\eta}| < \delta$, $\widehat{\phi}$  and $\pi_{u_1}$ are saved ($\phi_1 = \widehat{\phi}$), $\epsilon$ is reset to its initial value and a new $\widehat{\phi}$ is created. The gating system then uses the frozen $\phi_1$ to check if the new set of $\tau$ samples is novel. Only novel sets are forwarded to update the new $\widehat{\phi}$. The algorithm iterates and learns ($\pi_{u_2}^*$, $\phi_2$) corresponding to the next easily learnable observation stream. The algorithm terminates when all abstractions have been learned. The final result is $(\pi^*, \Phi)$. 

\textbf{Hyper Parameters.} The hyper parameters that are used by the algorithm are as follows: (1) IncSFA learning rate $\nu$, (2) threshold $\delta$, (3) $\beta$, (4) $\epsilon$ decay multiplier, (5)  $\tau$ and (6) reward parameter $\sigma$. $\nu$ is quite intuitive to set~\cite{kompella:nc2012}. $\delta$ is generally set to values between $0.0004-0.001$ depending on how well the expert modules need to encode the inputs. $\beta$ is set to $\nu\log 2/(2(n-1))$, where $n$ is the number of streams (a derivation is beyond the scope of this paper). See Section~\ref{SE:RESULTS_NONST}
for a discussion on setting $\tau$ and $\sigma$. A Python code of CD-MISFA 2.0 is available for download at \url{https://varunrajk.gitlab.io/}

\section{Experimental Results}
\label{SE:RESULTS}

Here, we evaluate the performance of our algorithm. The desired result is a sequence of SF abstractions acquired in the order of increasing learning difficulty. We use \emph{curiosity function} values~\cite{luciw2013intrinsic} as a metric to quantify the learning difficulty of an observation stream w.r.t IncSFA.  

\subsection{CD-MISFA 1.0 vs CD-MISFA 2.0}
\label{SE:RESULTS_COMP}

We compare our method with the previous CD-MISFA 1.0 algorithm: 

\textbf{(a)} CD-MISFA 1.0 uses a clustering algorithm called the Robust Online Clustering (ROC)~\cite{kaddc:2005} coupled to the IncSFA. ROC maintains estimates of IncSFA outputs that are correlated to some pre-defined discrete meta-class labels (\textit{e.\@g.\@} proprioception; the joint angles of a humanoid robot).  The ROC error is used to decide when to stop updating $\widehat{\phi}$ and to check if $\tau$ samples are novel. The disadvantages of using ROC are: (a)~It requires discrete meta-class labels, which can be hard to provide in general environments (\eg~see Section~\ref{SE:RESULTS_iCub}). (b)~It limits the abstractions to be correlated to the labels. (c)~It restricts the algorithm's application to discrete environments. (d)~It adds to the overall computational complexity. CD-MISFA 2.0 does not use ROC, instead it uses the low-complex, continuous-time slowness measure to check when to stop learning and how to filter the encoded inputs. This extends its application to continuous domains with relatively fewer hyper parameters to be set. The method does not require any meta-class labels and the abstractions learned are not constrained in any way.

 \begin{figure}[!t]
\includegraphics[width=0.99\linewidth]{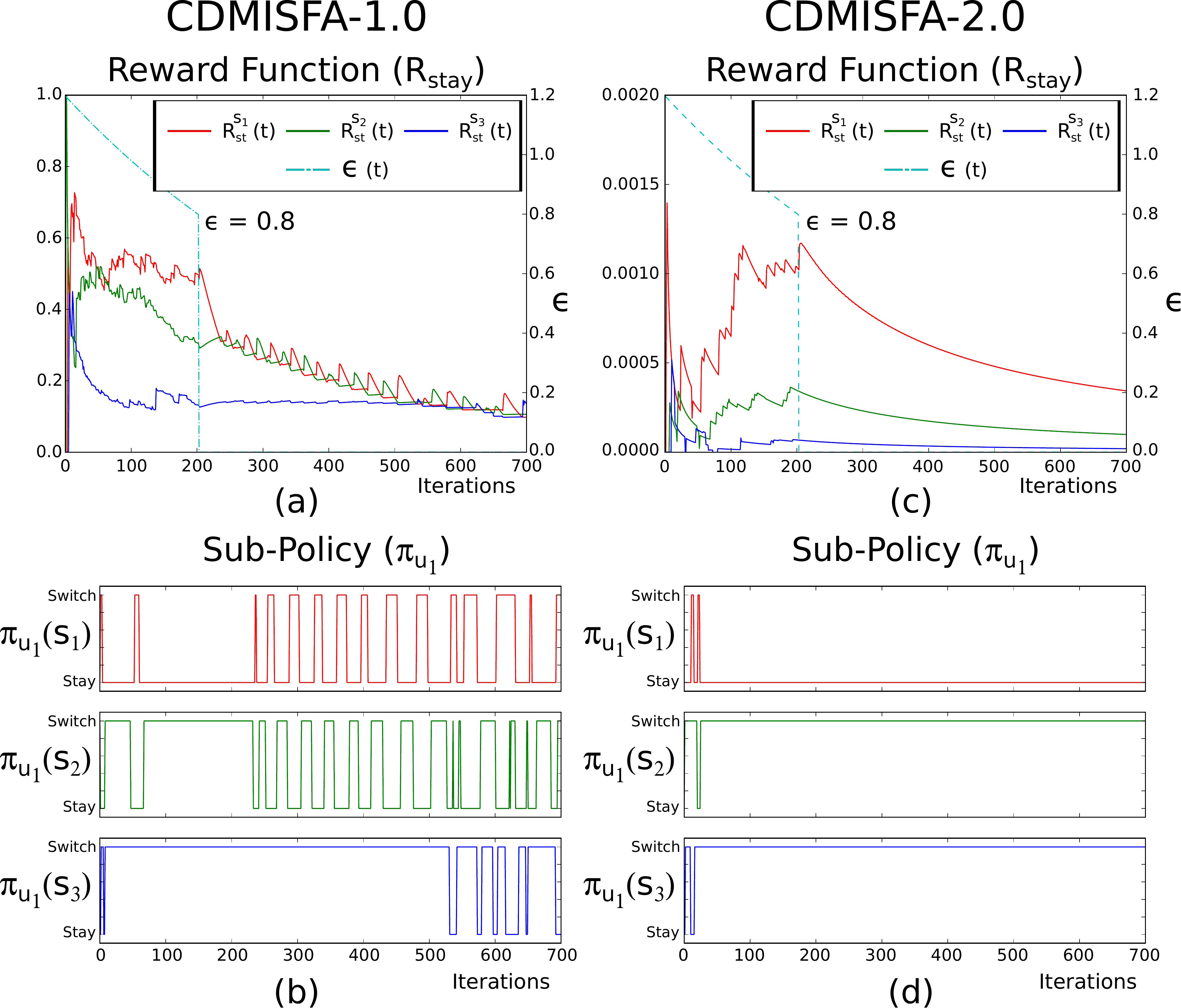}
\caption{\textbf{CD-MISFA 1.0 vs CD-MISFA 2.0.} (a) CD-MISFA 1.0 Reward Function (stay action only). It gets updated locally and this results in instability. (b) Unstable CD-MISFA 1.0 sub-policy $\pi_{u_1}$. (c) CD-MISFA 2.0 reward function and its stable (d) sub-policy $\pi_{u_1}$.}
\label{FIG:Compare}
\vspace{-0.3cm}
\end{figure}	

\textbf{(b)} CD-MISFA 1.0 uses a tabular reward function update rule~\cite{kompellaThesis}:
$\widetilde{R}^{ss'}_a   \leftarrow \alpha~ r^{ss'}_a +~(1-\alpha) \widetilde{R}^{ss'}_a;~
R \leftarrow \widetilde{R}/\|\widetilde{R}\|,$ where $\alpha$ is a constant. This rule only makes local tabular updates of the $(s,a,s')$ tuple entries. We found cases where CD-MISFA 1.0 becomes unstable using this reward function. To demonstrate this, we select an environment consisting of 3 nonlinear oscillatory streams~\cite{kompellaThesis} each learnable by IncSFA: 
\begin{eqnarray}
\label{EQ:S_T1}
 {\bf x}_1 &:& \left \{
\begin{array}{ll}
  x_{1}(t) = \sin(4~\theta_t-\pi/4.) - \cos(44~\theta_t)^2 \\
  x_{2}(t) = \cos(44~\theta_t)
\end{array},
 \right.\\
\label{EQ:S_T2}
 {\bf x}_2 &:& \left \{
\begin{array}{ll}
  x_{1}(t) = \sin(3~\theta_t) + \cos(27~\theta_t)^2 \\
  x_{2}(t) = \cos(27~\theta_t)
\end{array}, \text{ and} 
 \right.\\
\label{EQ:S_T3}
 {\bf x}_3 &:& \left \{
\begin{array}{ll}
  x_{1}(t) = \cos(12~\theta_t) \\
  x_{2}(t) = \cos(2~\theta_t) + \cos(12~\theta_t)^2
\end{array},
 \right.
\end{eqnarray}
where $\theta_t = 2\pi t/500$.
It can be found based on the learning difficulty values~\cite{kompellaThesis} that the slowest  feature of the stream ${\bf x}_1$ is the easiest to learn followed by ${\bf x}_2$ and then ${\bf x}_3$. The learning parameters are set as $\nu=0.05$, $\tau = 100$, $\sigma = 0.0009$. We initialized $\epsilon$ to $1.2$, so that the agent explores long enough. However, when used as a probability, any value of $\epsilon > 1$ is considered as $1$. $\epsilon$ decays with a multiplier equal to $0.998$ and is set to $0$ when it reaches the value of $0.8$. Figure~\ref{FIG:Compare}(a) shows the updating CD-MISFA 1.0 reward function for the \emph{stay} action over algorithm iterations. Since ${\bf x}_1$ is the easiest to learn, the algorithm finds the \emph{stay} action in $s_1$ most rewarding. As $\epsilon$ decays $<1$, the agent tends to spend more time in $s_1$ updating the reward function locally. When $\epsilon$ is set to zero, the reward function corresponding to $s_1$ decreases (because the curiosity rewards diminish), while the rest of the reward function entries remain the same. This results in an unstable policy as soon as $s_1$ reward value goes below that of $s_2$ and the module hasn't converged yet. The instability reoccurs for the reward value at $s_2$. This is not the case in CD-MISFA 2.0 (Figure~\ref{FIG:Compare}(c),(d)). The reward function is estimated using rewards that modify the whole function (Eq.~\eqref{EQ:REWARDUPDATE}). The policy therefore remains stable. 

\subsection{Oscillatory Streams Environment}
\label{SE:RESULTS_OSCI}

\begin{figure}[!t]
\includegraphics[width=\linewidth]{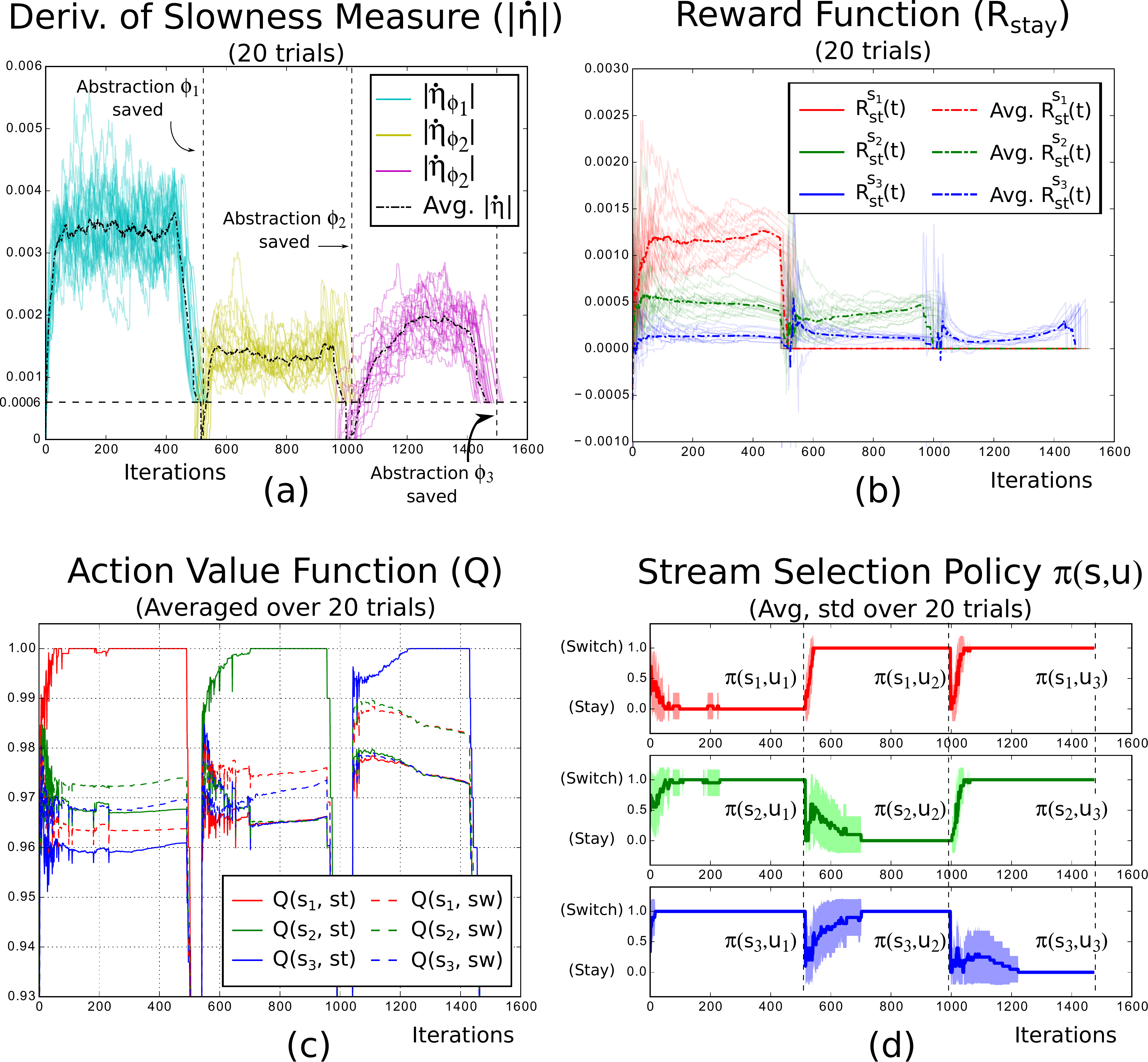}
\caption{\textbf{Oscillatory Streams Environment.} Experiment conducted with 20 trials of random initialization. (a) Derivative of the slowness measure over the algorithm iterations for the 20 trials. Dashed line indicates the average over all trials. A module is saved whenever the slowness measure drops below a threshold $\delta=0.0006$. (b) Reward function over iterations for the 20 trials. Stay action (\emph{st}) is state $s_1$ is most rewarding while learning the first module. Once the module is saved, stay action in state $s_2$ is most rewarding since inputs from $s_1$ are already encoded. The same is reflected in the (c) learned value function and the (d) stream selection policy. Figures are  best viewed in color.}
\label{FIG:full_RES}
\vspace{-0.3cm}
\end{figure}

We now investigate further the complete learning behavior of CD-MISFA 2.0 in the environment considered above. We used the same set of hyper-parameters: $\nu=0.05$, $\delta=0.0006$, $\tau=100$, $\sigma = 0.0009$, $\epsilon$ is initialized to $1.2$, with a $0.999$ decay multiplier. However, when $\epsilon<0.8$, the decay multiplier is set to $0.95$ to speed up the experiment. We executed the algorithm for 20 trials with different random initializations (seeds) and achieved optimal results for all the trials. An optimal result here is the abstraction set $\Phi^* = \{ \phi_1,\phi_2,\phi_3\}$, where $\phi_1$ encodes ${\bf x}_1$ (easiest to learn),  $\phi_2$ encodes ${\bf x}_2$ (next easier), and $\phi_3$ encodes ${\bf x}_3$. The optimal result also includes the policy to learn these abstractions for the given environment; $\pi^* = \{[0,1,1],[1,0,1],[1,1,0]\}$. 

Figure \ref{FIG:full_RES} shows the results of the experiment. For each trial, the agent begins exploring the three streams initially by executing actions \emph{stay} and \emph{switch} at random. The derivative of $\eta$ is high as the agent switches between the streams (Figure \ref{FIG:full_RES}(a)). During this period, $R$ becomes stable (Figure \ref{FIG:full_RES}(b)). Since  ${\bf x}_1$ is the easiest stream to encode, the \emph{stay} action in state $s_1$ is most rewarding. This is also reflected in the value function (averaged over 20 trials; Figure \ref{FIG:full_RES}(c)) and the sub-policy learned (Figure \ref{FIG:full_RES}(d)). As $\epsilon$ decays, the agent begins to exploit the learned sub-policy and the $\dot{\eta}$ begins to drop. Once it drops below $\delta$, the adaptive abstraction is saved $\phi_1 = \widehat{\phi}$ and a new $\widehat{\phi}$ is created. The process repeats, but now the gating system prevents re-learning ${\bf x}_1$ and therefore the agent finds staying in $s_2$ most rewarding. It learns an abstraction corresponding to ${\bf x}_2$ and then ${\bf x}_3$. This experiment has demonstrated that the algorithm learns the optimal policy in a stationary environment. 

\subsection{Non-Stationary Dynamic Environments} 
\label{SE:RESULTS_NONST}

\begin{figure}[!t]
\includegraphics[width=\linewidth]{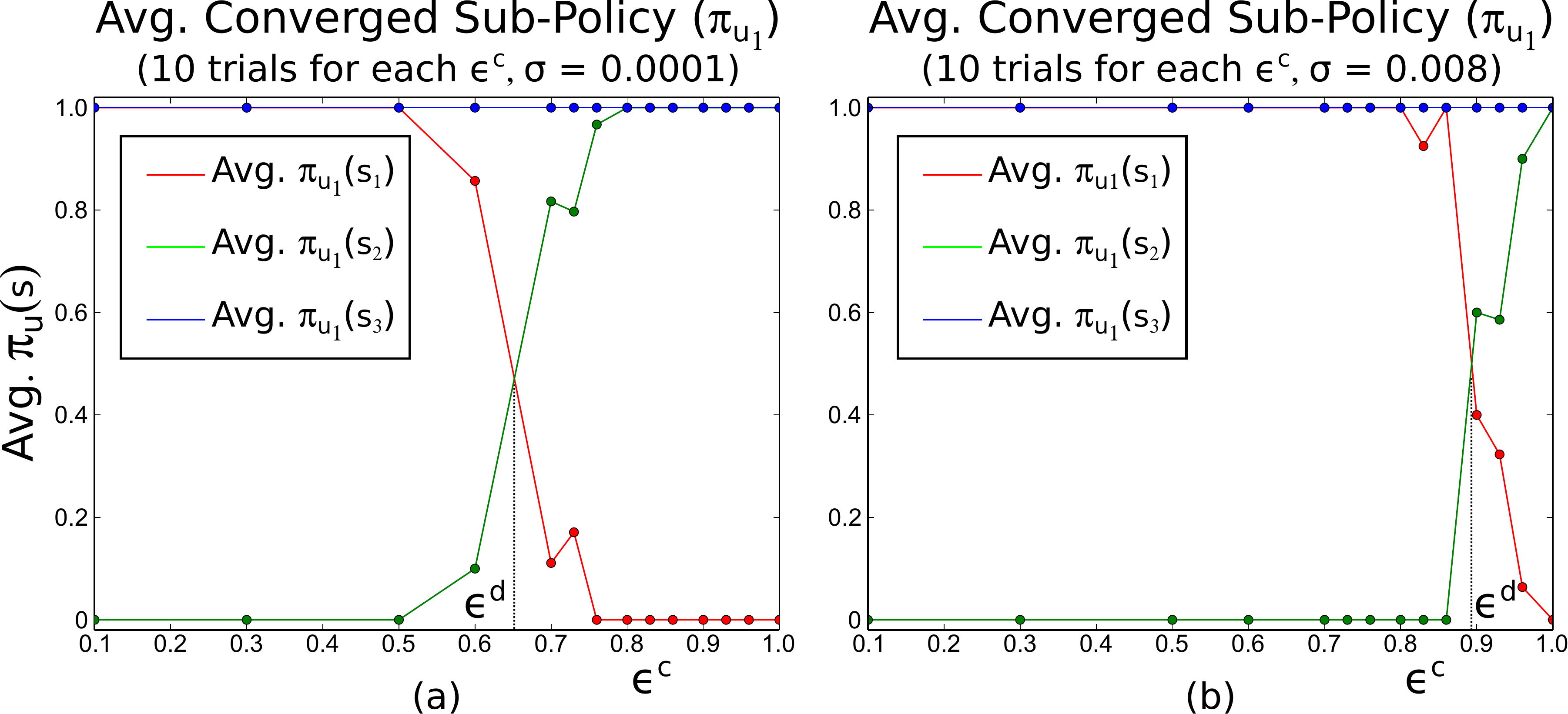}
\caption{\textbf{Non-Stationary Dynamic Environments.} Converged sub-policy averaged over 10 trials for each value of $\epsilon$ with (a) $\sigma = 0.0001$ and (b) $\sigma = 0.008$. For $\epsilon < \epsilon^d$, the policy converges to the old optimal with \emph{stay} (= 0) in $s_2$ and for $\epsilon > \epsilon^d$, the policy converges to the new optimal with \emph{stay} in $s_1$. Figures best viewed in color.}
\label{FIG:ED_RES}
\end{figure}	

Here, we discuss results of experiments conducted in non-stationary environments, where the  statistics changes abruptly in time. Consider an environment with 3 streams; the first stream is generating zeros, the second stream is ${\bf x}_2$ (Eq.~(\ref{EQ:S_T2})) and the third is ${\bf x}_3$ (Eq.~(\ref{EQ:S_T3})). Since ${\bf x}_2$ is easier to learn than ${\bf x}_3$, the optimal sub-policy is $[1,0,1]$. We let the algorithm's policy stabilize and when $\epsilon$ of the decaying $\epsilon$-greedy strategy falls below a constant $\epsilon^c$, we replace the zero-stream with ${\bf x}_1$ (Eq.~(\ref{EQ:S_T1})). The new optimal sub-policy after that signal swap is $[0,1,1]$, since ${\bf x}_1$ is now the easiest to learn. For the rest of this section, we denote $[1,0,1]$ as the old optimal sub-policy and $[0,1,1]$ as the new optimal sub-policy. We simulate different non-stationary environments by setting different values for $\epsilon^c \in \{$ 1., 0.96, 0.93,  0.9,  0.86, 0.83,  0.8, 0.76, 0.73, 0.7, 0.6, 0.5, 0.3, 0.1$\}$.  For these non-stationary environments, we address the following questions:
\begin{enumerate}
 \item Is the algorithm stable when $\epsilon$ decays to zero? That is, does it converge to a particular policy consistently over many trials of random initializations?
 \item To which policy does the algorithm converge? 
 \item What hyper-parameters effect the result?
\end{enumerate}

First, we discuss the performance of the algorithm for hyper-parameters similar those in the previous experiments, except for $\sigma=0.0001$. Figure~\ref{FIG:ED_RES}(a) shows the learned sub-policy $\pi_{u_1}$ (after $\epsilon \approx 0$) averaged over 10 randomly initialized (seed) trials for each value of $\epsilon^c \in \{$ 1., 0.96, 0.93,  0.9,  0.86, 0.83,  0.8, 0.76, 0.73, 0.7, 0.6, 0.5, 0.3, 0.1$\}$. It is clear that there is a value $\epsilon^d\leq1$, so that for $\epsilon^c > \epsilon^d$, the algorithm consistently converges to the new optimal policy (except for the values close to $\epsilon^d$). While, for $\epsilon^c < \epsilon^d$ the algorithm converges to the old optimal policy. We denote $\epsilon^d$ as the point-of-no-return $\epsilon$. This result shows that the algorithm remains stable when the $\epsilon$ decays to zero, and converges to the old optimal policy if the environment statistics change at any $\epsilon < \epsilon^d$.  If the environment changes when $\epsilon > \epsilon^d$, then the algorithm learns the new optimal policy consistently. Next, we discuss if different hyper parameters effect this behavior. 

\begin{table}[!t]
\centering
 \begin{tabular}{|c||c|c|c|c|c|c|}
 \hline
  $\sigma$ & 0.008 &  0.003 & 0.0009 &  0.0001 & 0\\
  \hline
  $\epsilon^d$ & 0.8933 &  0.8775 & 0.7211  & 0.6517 & 0.6483\\
  \hline
 \end{tabular}
\caption{$\epsilon^d$ vs $\sigma$ (10 randomly initialized trials for each $\sigma$)}
\label{TAB:Comp1}
\end{table}

\begin{table}[!t]
\centering
 \begin{tabular}{|c||c|c|c|c|}
  \hline
  $\nu$ & 0.02 & 0.03 & 0.04 & 0.05\\
  \hline
  $\epsilon^d$ & 0.78 & 0.80 & 0.79 & 0.81\\
  \hline
 \end{tabular}
 \begin{tabular}{|c||c|c|c|c|}
 \hline
  $\tau$ & 10 & 30 & 50 & 100\\
  \hline
  $\epsilon^d$ & 0.98 & 0.81 & 0.83 & 0.80\\
  \hline
 \end{tabular}
\caption{$\epsilon^d$ vs $\nu$, $\tau$ (10 randomly init.\ trials for each $\nu$, $\tau$)}
\label{TAB:Comp2}
\end{table}

Figure~\ref{FIG:ED_RES}(b) shows the same experiment with 10 random initializations for a higher value of $\sigma=0.008$. It is clear that for a higher value of $\sigma$, $\epsilon^d$ is higher,  therefore, pushing the decision boundary to stick to the old optimal. This is also evident from the Table~\ref{TAB:Comp1}. $\sigma$ controls the effect of the expert rewards (Eq.~\eqref{EQ:REWPROG}). Therefore, expert rewards bias the agent to become an expert by exploiting the learned old optimal instead of exploring to learn the new optimal. 
Lastly, we have also conducted the same experiment for different values of IncSFA learning rate $\nu$ and $\tau$, keeping the rest of the parameters fixed to their values of the previous experiment. Table~\ref{TAB:Comp2} shows how $\nu$ and $\tau$ have no significant effect on $\epsilon^d$, with the exception of $\tau=10$, where we suspect the value is too low to estimate the rewards properly. The above results show that the algorithm is stable in the above non-stationary environments and converges to either the old optimal or the new optimal solution depending on the value of $\epsilon$. The result also demonstrates the effect of the expert rewards on the system. 

\subsection{Curiosity-Driven Vision-Enabled iCub} 
\label{SE:RESULTS_iCub}

\begin{figure}[!t]
\includegraphics[width=\linewidth]{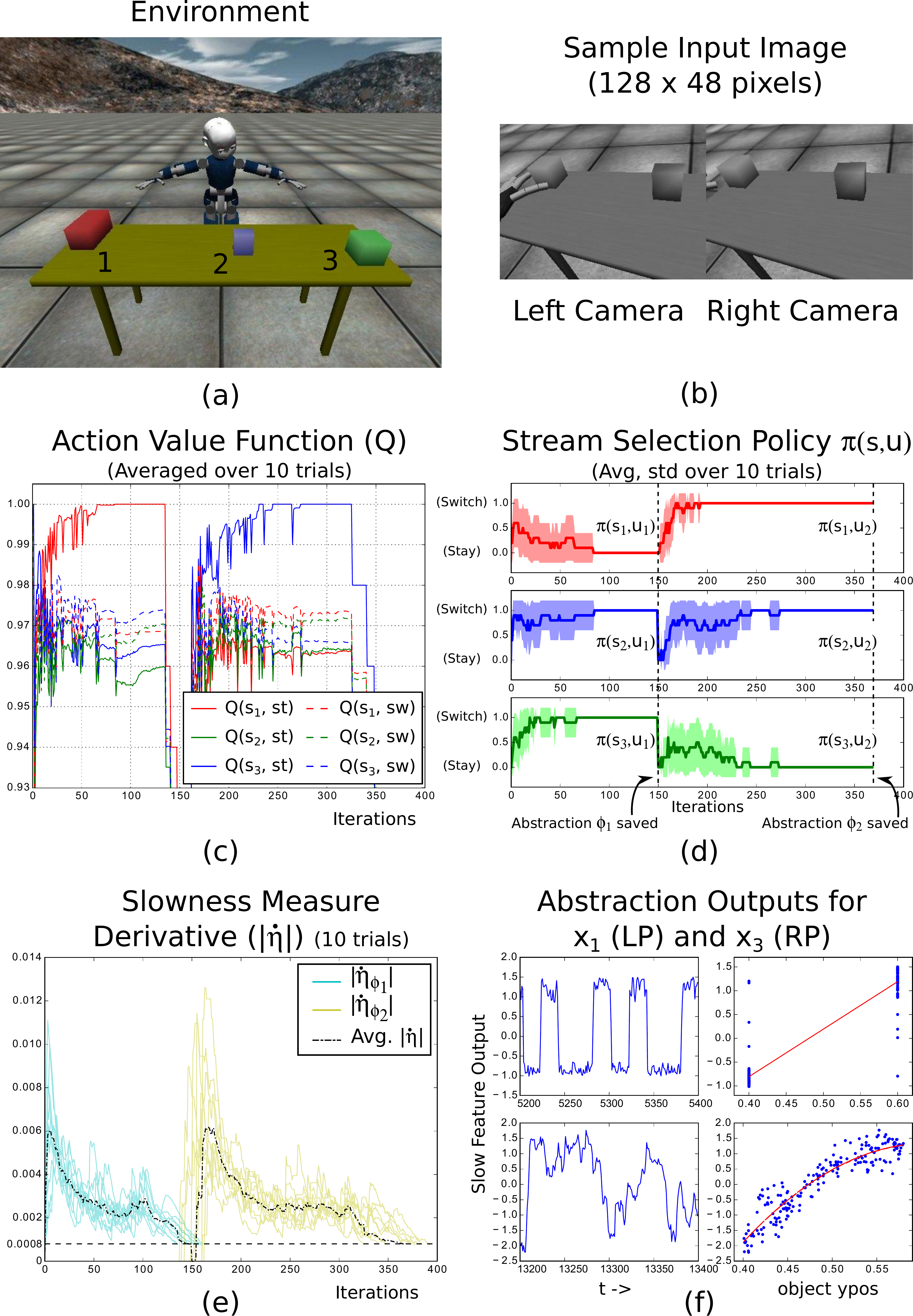}
\caption{\textbf{Curiosity-Driven Vision-Enabled iCub.} Experiment conducted with 10 trials of random initialization in the iCub Simulator. (a) The environment consists of an iCub placed next to a table with three moving objects. The iCub has a limited field of view and can rotate its head over three perspectives $\{s_1, s_2, s_3\}$ to observe the objects. It receives continuous streams of image observations through its camera-eyes. (b) A sample observation. (c) Averaged action value function over time. The iCub finds object 1 dynamics most interesting to learn followed by object 3 and finds object 2's unlearnable dynamics un-interesting. (d) Average and std. deviation (shaded region) of the stream selection policy: $\{[0,1,1],[1,1,0]\}$. (e) Derivative of the slowness measure over the algorithm iterations for the 20 trials. Dashed line indicates the average over all trials. A module is saved whenever $|\dot{\eta}| < \delta=0.0006$. (f) Outputs of abstractions learned. Both abstractions encode object 1\&3's positions. See text for details. Figures are  best viewed in color.}
\label{FIG:iCUB_RES}
\vspace{-0.3cm}
\end{figure}

An important open problem in vision-based developmental robotics is, how can an online vision-enabled humanoid robot akin to a human baby focus/shift its attention towards events that it finds interesting? Can its curiosity to explore also drive learning abstractions? We present here an experiment to demonstrate that this is possible using CD-MISFA 2.0. To this end, we use the iCub Simulation software~\cite{iCubSim}. An iCub is placed next to a table with three objects of different sizes (Figure~\ref{FIG:iCUB_RES}(a)). The environment is dynamic and continuous; all the three object's positions (unknown to the iCub) change at every time $t$. Object-1's x-position changes uniformly randomly within the range (-0.4,-0.6) and its y-position is either 0.4 or 0.6 and toggles at a fixed unknown frequency. Both x and y-position of object-2 change uniformly randomly. Object-3 performs a random walk with its y-position changing slowly compared to its x-position. The three object's movements depict three distinct dynamic events in the iCub's environment. 

The iCub has two onboard camera eyes and the images captured are converted to grayscale and downscaled to a size of 128x48 pixels. Figure~\ref{FIG:iCUB_RES}(b) shows a sample input image. The iCub explores by rotating its head over a single joint. We use three joint positions such that it can view the objects over three overlapping perspectives: left (LP), center (CP) and right (RP), each generating a stream of high-dimensional observations $\{{\bf x}_1, {\bf x}_2, {\bf x}_3\}$. IncSFA finds the streams ${\bf x}_1$ and ${\bf x}_3$ learnable and ${\bf x}_2$ unlearnable since only object-1 and 3's positions have a temporal structure. Furthermore, we calculated the learning difficulty values~\cite{luciw2013intrinsic} and found that ${\bf x}_1$ is easier to encode by IncSFA than ${\bf x}_3$. 

It is not straightforward to apply CD-MISFA 1.0 in this environment since the dynamics (changing object's positions) have no correlation to the robot's proprioception. Therefore, it is hard to provide any discrete meta-class labels to the ROC (see Section~\ref{SE:RESULTS_COMP}) to make any  progress in learning abstractions. On the other hand, since CD-MISFA 2.0 does not require any pre-defined labels, we expect that it first learns an abstraction encoding the position of object-1 and then an abstraction encoding the position of object-3 (see Kompella \textit{et al.\@}'s work~\cite{kompella:nc2012} for details on why IncSFA learns the positions). The experiment would then terminate after this as there are no other IncSFA learnable events in the environment. 

We used hyper-parameters similar to the previous experiments, except for $\nu = 0.01$, $\tau = 40$, $\sigma = 0.01$, $\delta = 0.0008$. $\cS = \{s_1, s_2, s_3\}$ corresponds to $\{{\bf x}_1, {\bf x}_2, {\bf x}_3\}$. We conducted 10 trials of the experiment with different random seed values and the algorithm found the optimal policy in all the trials. Figures~\ref{FIG:iCUB_RES}(c)-(f) show the cumulative results. For each trial, the iCub starts exploring by moving its head using the actions \emph{stay} and \emph{switch}. It receives high curiosity-rewards for the observations from ${\bf x}_1$ compared to the other streams. Therefore, as the $\epsilon$ decays, it finds the \emph{stay} action in state $s_1$ most valuable (Figure~\ref{FIG:iCUB_RES}(c)) and the sub-policy converges to $\pi_{u_1} = [0,1,1]$ (Figure~\ref{FIG:iCUB_RES}(d)). The converging $\pi_{u_1}$ enables $\widehat{\phi}$ to converge and $|\dot{\eta}|$ begins to drop (Figure~\ref{FIG:iCUB_RES}(e)). Once it drops below $\delta$, the adaptive abstraction is saved ($\phi_1 \leftarrow \widehat{\phi}$), $\epsilon$ is reset and a new $\widehat{\phi}$ is created. The process repeats, but the gating system prevents re-learning ${\bf x}_1$ and the agent now learns $\pi_{u_2} = [1,1,0]$ and an abstraction $\phi_2$ corresponding to ${\bf x}_3$. The process continues, however, the system never converges to a third abstraction since the dynamics of ${\bf x}_2$ is uniformly random (therefore not shown in the figures). Figure~\ref{FIG:iCUB_RES}(f) top-left shows the output of ${\bf y}(t) = \phi_1 {\bf x}_1(t)$. $\phi_1$ encodes the two y-positions of object-1. This is also evident from Figure~\ref{FIG:iCUB_RES}(f) top-right, where we plotted the last 200 output values (before the abstraction was frozen) with respect to the y-position of object-1. The red line shows a polynomial fit over these values. Similarly, Figures~\ref{FIG:iCUB_RES}(f) bottom show that $\phi_2$ encodes the y-position of object-3. How can these abstractions be useful? They can be used by the iCub to interact with the objects in a predictable way. An eight times sped up video of this experiment can be found here: \url{https://varunrajk.gitlab.io/videos/iCubExp8x.mp4} 

\section{Conclusion}
\label{SE:CONC}

This paper presents an online learning system that enables an agent to learn to look in regions where it can find the next easiest yet unknown regularity in its high-dimensional sensory inputs. We have shown through experiments that the method is stable in certain non-stationary environments. The iCub experiment demonstrates that the reliable performance of the algorithm extends to high-dimensional image inputs, making it valuable for vision-based developmental learning. Our future work involves implementing the algorithm in environments where the input observation streams are generated as a consequence of executing different time-varying behaviors (e.g.\  \emph{options}~\cite{sutton1999between}) and also in environments where it can learn to reuse the learned modular abstractions to solve an external task.

\bibliographystyle{plain}
\bibliography{/home/kompella/Dropbox/Publications/master}

\end{document}